\pdfoutput=1

\documentclass[11pt]{article}
\usepackage[table]{xcolor}
\usepackage{tcolorbox}
 \usepackage{microtype}
\usepackage[table]{xcolor}
\usepackage{booktabs}
\usepackage{amssymb}
\newcommand{\cmark}{\ding{51}}%
\newcommand{\xmark}{\ding{55}}%
\usepackage{pifont}
\usepackage{makecell}
\usepackage[final]{acl}
\usepackage{epigraph}
\usepackage{dirtytalk}
\usepackage{times}
\usepackage{latexsym}
\definecolor{Salmon}{rgb}{0.98, 0.5, 0.45}
\definecolor{YellowGreen}{rgb}{0.6, 0.8, 0.2} 
\usepackage{listings}

\definecolor{codegreen}{rgb}{0,0.6,0}
\definecolor{codegray}{rgb}{0.5,0.5,0.5}
\definecolor{codepurple}{rgb}{0.58,0,0.82}
\definecolor{backcolour}{rgb}{0.95,0.95,0.92}

\lstdefinestyle{mystyle}{
    backgroundcolor=\color{backcolour},   
    commentstyle=\color{codegreen},
    keywordstyle=\color{magenta},
    numberstyle=\tiny\color{codegray},
    stringstyle=\color{codepurple},
    basicstyle=\ttfamily\footnotesize,
    breakatwhitespace=false,         
    breaklines=true,                 
    captionpos=b,                    
    keepspaces=true,                 
    numbers=left,                    
    numbersep=5pt,                  
    showspaces=false,                
    showstringspaces=false,
    showtabs=false,                  
    tabsize=2
}

\usepackage{cuted}
\usepackage{listings}
\definecolor{bggray}{rgb}{0.95,0.95,0.95}
\usepackage[T1]{fontenc}

\usepackage[utf8]{inputenc}

\usepackage{microtype}

\usepackage{inconsolata}

\usepackage{graphicx}

\title{MOLE: Metadata Extraction and Validation in Scientific Papers Using LLMs}

\author{
Zaid Alyafeai\textsuperscript{1} \,\,\,\,
Maged S. Al-Shaibani\textsuperscript{2} \,\,\,\,
Bernard Ghanem\textsuperscript{1} \\
\textsuperscript{1}KAUST \quad
\textsuperscript{2}SDAIA-KFUPM Joint Research Center for AI, KFUPM
}

\begin{document}
\maketitle

\begin{abstract}
Metadata extraction is essential for cataloging and preserving datasets, enabling effective research discovery and reproducibility, especially given the current exponential growth in scientific research. While Masader \cite{alyafeai2021masader} laid the groundwork for extracting a wide range of metadata attributes from Arabic NLP datasets' scholarly articles, it relies heavily on manual annotation. In this paper, we present MOLE, a framework that leverages Large Language Models (LLMs) to automatically extract metadata attributes from scientific papers covering datasets of languages other than Arabic. Our schema-driven methodology processes entire documents across multiple input formats and incorporates robust validation mechanisms for consistent output. Additionally, we introduce a new benchmark to evaluate the research progress on this task. Through systematic analysis of context length, few-shot learning, and web browsing integration, we demonstrate that modern LLMs show promising results in automating this task, highlighting the need for further future work improvements to ensure consistent and reliable performance. We release the code\footnote{\url{https://github.com/IVUL-KAUST/MOLE}} and dataset\footnote{\url{https://huggingface.co/datasets/IVUL-KAUST/MOLE}} for the research community. 
\end{abstract}

\section{Introduction}
\,\,\, \,\,\, \,\,\, \,\,\, \say{\textit{Metadata is data about data}}

\begin{figure}[!htp]
    \centering
    \includegraphics[width=0.7\linewidth]{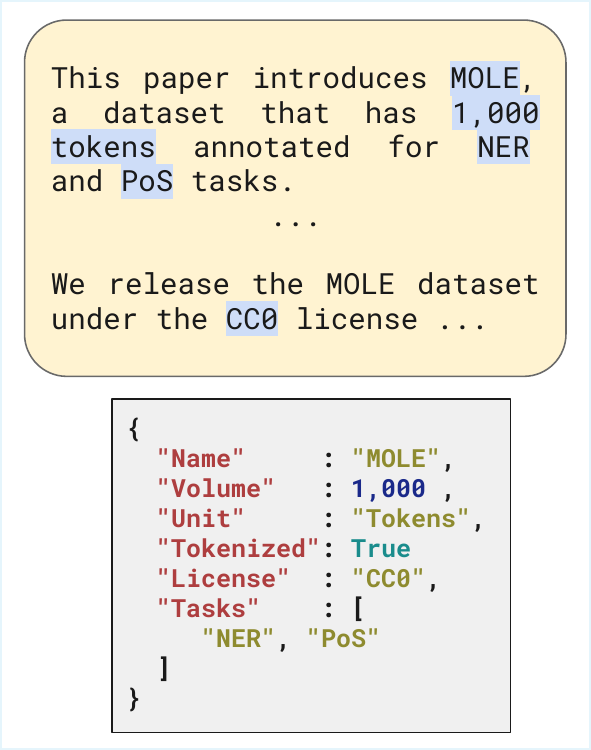}
    \caption{Sample metadata extracted from a dummy paper with highlighted attributes.}
    \label{fig:main}
\end{figure}

The scientific community is experiencing an unprecedented data revolution, with researchers producing and sharing datasets at an extraordinary rate. However, the value of these datasets decreases when they are inadequately documented or difficult to discover. Extracting structured information that describes the characteristics, origins, and usage of datasets, is a critical challenge \cite{borgman2012conundrum, wilkinson2016fair}, especially in vastly and rapidly growing domains such as natural language processing (NLP). These datasets vary widely in structure, size, format, purpose, and language. Without robust metadata extraction, valuable datasets remain underutilized, research efforts are duplicated, and reproducibility is compromised \cite{gebru2021datasheets, dodge2019show}. With hundreds of thousands of new publications annually\footnote{As of April 2025, there are more than 2.7 million articles on \href{https://arxiv.org/stats/monthly\_submissions}{arXiv}.}, automating metadata extraction is essential to maintain the scalability of the scientific ecosystem.

\begin{table*}[]
\centering
\small
\caption{Comparison between MOLE and other methods in the literature.}
\label{tab:lit}
\begin{tabular}{p{4cm}ccccccc}
\toprule
\textbf{Method}                                                         & \textbf{Attributes} & \textbf{Models}   & \textbf{Schema} & \textbf{Benchmark} & \textbf{Formats}           & \textbf{Browsing}  \\
\midrule
MOLE                                                          & {32}            & 7 LLMs     & \cmark           & {126}                  & {3} & \cmark              \\
\cite{watanabe2024capabilities}                                     & 8             & 2 LLMs     & 	\xmark            & \xmark                         & 1                    & \xmark               \\
\cite{giner2022describeml} & 23            & 2 LLMs     & \xmark            & 12                   & 1                    & \xmark               \\
\cite{ahmad2020flag}                            & 9             & SVC        & \xmark            & \xmark                         & 1                    & \xmark               \\
\cite{tkaczyk2015cermine}                                & 17            & SVM        & \xmark            & \xmark                         & 1                    & \xmark               \\
\cite{constantin2013pdfx}                                & 18            & Rule-based & \xmark            & \xmark                         &           1          & \xmark               \\
\cite{councill2008parscit}                               & 23            & CRF        & \xmark            & 40                         &        1             & \xmark              \\
\bottomrule
\end{tabular}
\end{table*}

This work attempts to approach this problem utilizing LLMs to extract the metadata. We define metadata as a JSON object that holds many attributes, such as Year, License, and Paper and Dataset Links. Such attributes vary in constraints: some fixed options (License), some free form (Dataset Description), and some context-dependent (Paper \& Dataset Links). While existing automated approaches typically extract around 5-10 attributes \cite{ahmad2020flag, tkaczyk2015cermine}, our work automatically extracts around 30 different attributes per paper, providing a substantially more comprehensive metadata profile (see Table \ref{tab:lit}). Figure \ref{fig:main} overviews our work of extracting sample attributes from a paper as a simplified example.

Current metadata extraction approaches typically rely on rule-based systems, supervised machine learning, or combinations thereof \cite{rodriguez2021mel,ahmad2020flag,tkaczyk2015cermine,constantin2013pdfx,granitzer2012comparison,councill2008parscit}. While effective for structured documents, these methods struggle with the heterogeneity of scientific papers and require domain knowledge and maintenance to accommodate evolving document structures \cite{rizvi2020hybrid}.

Recent advances in Large Language Models (LLMs) have opened new possibilities for information extraction \cite{team2023gemini, openai2023gpt4}, with models showing promising results in extracting structured data, \cite{butcher2025precise,li2024simple,liu2024we,tam2024let}. A prominent advantage of modern LLMs is their ability to handle long contexts \cite{peng2023yarn, zhang2023extending, team2023gemini}, allowing our methodology to process entire papers. We summarize our contributions as follows:

\begin{figure*}
\centering
\includegraphics[width=0.9\linewidth]{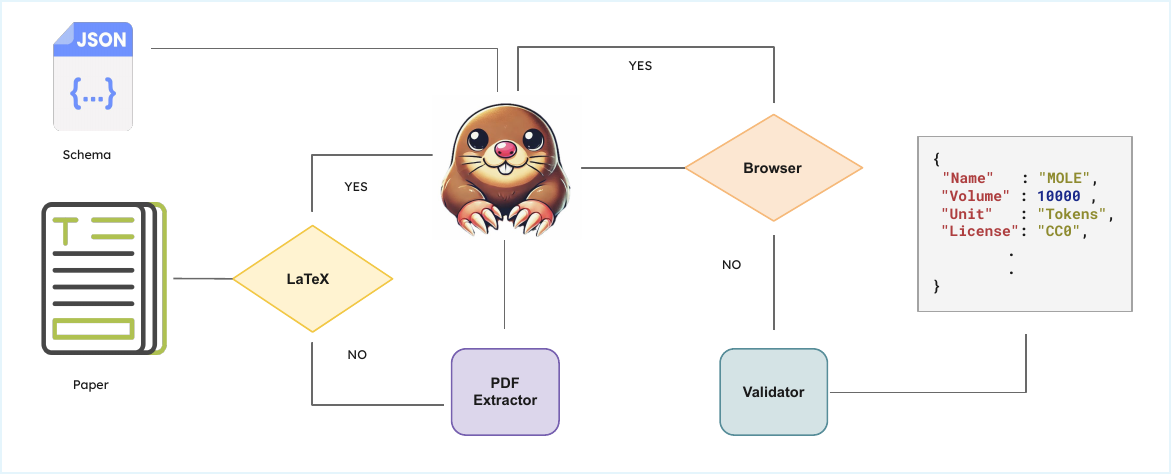}
\caption{MOLE pipeline. The paper text and Schema are used as input, and the output is the extracted metadata content. }
\label{fig}
\end{figure*}
\begin{enumerate}
    \item A generalized approach for dataset metadata extraction from scientific papers, capable of extracting more than 30 distinct attributes organized in a structured schema hierarchy. 
    \item A benchmark for metadata extraction involving datasets in multiple languages, covering Arabic, English, Russian, French, and Japanese, enabling systematic evaluation of performance across linguistic domains.
    \item An examination of our approach on 7 LLMs, including proprietary and open-source models, analyzing the impact of long-context handling, few-shot learning, and constrained output generation.
\end{enumerate}

\section{Methodology}

Figure \ref{fig} illustrates our MOLE framework. This framework processes scientific papers in either LaTeX source or PDF format, leveraging the power of LLMs' extraction capabilities to identify dataset metadata attributes. When processing LaTeX, the framework directly analyzes the source; for PDF, we study extracting text manually using the available PDF tools vs. prompting an LLM (with vision capabilities) to extract a structured output from the paper. Then, an LLM identifies and structures the metadata according to our schema, which is subsequently validated before producing the final JSON output. This pipeline enables an automated, efficient, and reliable extraction of a comprehensive dataset metadata from diverse document formats and across multiple languages.
\subsection{Metdata}
We build on the work of Masader \cite{alyafeai2021masader} by modifying the following attributes:

\begin{itemize}
    \item \textbf{HF Link} we add the Hugging Face\footnote{Currently, \url{https://huggingface.co/datasets} contains more than 300K datasets. } link to the dataset.
    \item \textbf{Access} we change the \textit{Cost} attribute to show the actual cost of the dataset and add \textit{Access} to highlight the accessibility of the dataset, including Free, Upon-Request, or Paid. 
    \item \textbf{Derived From} we replace \textit{Related datasets} by this attribute to mention all datasets that are used as a seed for a given dataset. This attribute is critical for evaluation as it can indicate any contamination issues.
    \item \textbf{Domain} we use the following options to describe the attribute: social media, news articles, commentary, books, Wikipedia, web pages, public datasets, TV channels, captions, LLM, or other. The options are improved by including recent approaches for synthetic data generation using LLMs.
    \item \textbf{Collection Style} similar to \textit{Domain} we use the following options to describe the attribute: crawling, human annotation, machine annotation, manual curation, LLM generated, or other. 
\end{itemize}

In total, we have 32 attributes that can be used to annotate Arabic datasets (See Figure \ref{fig:groups}). We also extend our approach to datasets in other languages by considering the more general attributes. In this study, we consider the following language categories: Arabic (ar), English (en), French (fr), Japanese (jp), Russian (ru), and multilingual (multi). Subsequently, we extend the metadata attributes to such categories. For example, the \textit{Script} attribute in Japanese could be  Kanji, Hiragana, Katakana, or mixed. In other languages, it is fixed, as it is Latin in English and French and Cyrillic in Russian. Note that other languages don't have dialects, so we remove the \textit{Dialect} and \textit{Subset} attributes for monolingual datasets. For the multilingual category, we remove the \textit{Script} attribute and use the \textit{Subsets} to indicate the languages in the dataset and their corresponding size.

\subsection{Schema}
We define a schema as a JSON representing how an LLM should generate the metadata. Each metadata attribute is represented by a key in the schema. Our metadata schema mainly consists of five keys:

\begin{enumerate}
    \item \textbf{question} specifies what metadata attribute to extract from the document. For example, we ask \textit{What is the license of the dataset?} for the \textit{License} attribute.
    \item \textbf{options} a list of string values to choose from. The LLM must choose an answer from this list. 
    \item \textbf{option description} a dictionary that explains ambiguous options, for example, it might not be clear what low, medium, or high Ethical Risks mean. 
    \item \textbf{answer type} this field represents the output type for each metadata attribute. The complete list of output types is shown in Table \ref{tab:types}.
    \begin{table}[!htp]
    \centering
    \small
    \caption{Permissible data types in our schema. The data types are provided in the schema to force the model to generate specific data types. }
    \label{tab:types}
    \begin{tabular}{ll} \toprule
         \textbf{Type} & \textbf{ Description}  \\ \midrule
         \texttt{str} & string \\
         \texttt{url} & link \\
         \texttt{date[year]} & year of the date \\
         \texttt{List[str]} & list of strings \\
         \texttt{float} & floating point number  \\
         \texttt{bool} & true or false \\ 
         \texttt{List[Dict]} & list of dictionaries  \\ \toprule
         
    \end{tabular}
\end{table}
    \item \textbf{validation group} this field is used to collect similar attributes in a group. Mainly we use this for evaluation.  
    \item \textbf{answer min and max} this field specifies the length of an answer for a given question. As an example, take the \textit{Tasks} attribute; then we assume that each dataset must have at least one task associated with it and at maximum 3 tasks. Hence we will have \textit{answer\_min = 1, answer\_max = 3}. In general, if \textit{answer\_min = 0}, this attribute is optional. If the \textit{answer\_max} is not defined, then there are no constraints on the output max length.  
    
\end{enumerate}


In the following example, we show a schema for the \textit{License} attribute. The answer min and max are set to 1 because a dataset must have only one license. 

\begin{lstlisting}[language=Python, caption=Example Schema for the License metadata attribute. Options are truncated for better visualization, numbers=none]
{
    "question": "What is the license of the dataset?",
    "options": [
        "Apache-2.0", "MIT", ...
    ],
    "answer_type": "str",
    "validation_group":"ACCESSIBILITY",
    "answer_min": 1,
    "answer_max": 1
}
\end{lstlisting}

\begin{table}[!htp]
    \centering
    \small
    \caption{Number of annotated papers and annotated metadata attributes for each category in the collected test dataset. }
    \label{tab:datasets}
    \begin{tabular}{lccc} \toprule
    \textbf{Category} & \textbf{\# papers} & \textbf{\# fields} & \textbf{\# annotations}  \\ \midrule
    ar & 21 & 64 & 1,344 \\
    en & 21 & 58 & 1,218 \\
    fr & 21 & 58 & 1,218 \\
    jp & 21 & 60 & 1,260 \\
    ru & 21 & 58 & 1,218 \\
    multi & 21 & 60 & 1,260 \\ \midrule
    total & 126 & 358 & 7,518 \\ \toprule
\end{tabular}
\end{table}

\begin{table*}[]
\centering
\small
\caption{Results of all models on the main categories Arabic (ar), English (en), Japanese (jp), French (fr), Russian (ru), Multilingual (multi) datasets. Average shows the weighted average of all categories. Maximum across category is \textbf{bold} and the second maximum is \underline{underlined}.}
\label{tab:main}
\begin{tabular}{lccccccc}
\toprule
\textbf{Model} & \textbf{ar} & \textbf{en} & \textbf{jp} & \textbf{fr} & \textbf{ru} & \textbf{multi} & \textbf{Average} \\
\midrule
\textbf{Random} & 32.29 & 27.86 & 32.74 & 30.27 & 32.13 & 23.18 & 29.74 \\
\textbf{Keyword} & 45.37 & 43.89 & 44.72 & 45.81 & 46.21 & 37.36 & 43.89 \\
\textbf{Gemma 3 27B} & 59.08 & 69.35 & 70.93 & 69.29 & 67.88 & 60.81 & 66.23 \\
\textbf{Qwen 2.5 72B} & 66.26 & 67.37 & 70.05 & 71.94 & 65.78 & 65.88 & 67.88 \\
\textbf{Llama 4 Maverick} & 60.77 & 73.38 & \underline{72.35} & 70.01 & 71.30 & 68.02 & 69.30 \\
\textbf{DeepSeek V3} & 66.64 & 73.78 & 71.26 & 72.13 & 71.48 & 64.88 & 70.03 \\
\textbf{Claude 3.5 Sonnet} & 65.50 & 71.14 & 71.63 & \textbf{75.71} & \underline{75.62} & \underline{68.45} & 71.34 \\
\textbf{GPT 4o} & \underline{67.32} & \underline{76.14} & 71.00 & 72.95 & 73.85 & 67.00 & \underline{71.38} \\
\textbf{Gemini 2.5 Pro} & \textbf{68.73} & \textbf{80.91} & \textbf{77.60} & \underline{75.06} & \textbf{78.00} & \textbf{71.09} & \textbf{75.23} \\
\bottomrule
\end{tabular}
\end{table*}

\subsection{Validation}
 We mainly use three types of validations to make sure the output is consistent with our schema:

\begin{enumerate}
    \item \textbf{Type Validation} if the output type for a given question is not correct, then we either cast it or use the default value. For example, the volume can be casted to \texttt{float} if it is given as \texttt{str}. 

    \item \textbf{Option Validation} if there are options to answer the question, then if the answer does not belong to one of the options, we use similarity matching to choose the most similar option.

    \item \textbf{Length Validation} the output length must be within the range [answer\_min, answer\_max], otherwise the model will have a low score for length enforcing. 

    \item \textbf{JSON Validation} The generated JSON must be loadable using \texttt{json.loads(...)}. To fix unloadable strings, we apply some regex rules. As an example, we remove \texttt{```json} prefixes in some generated JSONs. 
\end{enumerate}

\subsection{Metrics}
We use either exact match or list matching, depending on the answer type in the schema. For list matching, we use set intersection to compare the results. We use flexible matching where we allow at most one difference between the predicted and the ground truth. For dictionary matching, in addition to values, we also match the keys. We use F1 to calculate the score for a given metadata extraction. The precision calculates how much LLMs hallucinate values of non-existent attributes in the paper (lower precision), while recall calculates the accuracy of predicting metadata that exists in the paper.

\section{Dataset}

We manually annotated 126 papers covering datasets in different languages for testing. We additionally annotate 6 more papers for validation. Two authors participated in the annotation process \footnote{Guidelines: \url{https://github.com/IVUL-KAUST/MOLE/blob/molev2_anonymized/GUIDLINES.md}}. We annotate each metadata with two values; the first one is the value of the metadata, and the second with a binary value of 1 if the attribute exists in the paper; otherwise, it is 0. The binary annotation will help us in measuring which metadata exists in the papers and which ones require browsing the Web. For example, the License of a given dataset might not exist in the paper itself, but most likely it can be accessed through the Link. The articles collected span six different categories: Arabic (ar), English (en), French (fr), Japanese (jp), Russian (ru), and Multilingual (multi) datasets. The annotation process included annotating a subset and calculating the agreement percentage of the two annotations. The percentage of agreement between the two authors was found to be 85.93. We will add the annotation guidelines to the paper.
In Table \ref{tab:datasets}, we highlight the annotated papers in each category in addition to the number of annotated metadata attributes in each category. Additionally, we create six schemata for the different language categories.
\begin{table}[!htp]
    \centering
    \small
    \caption{Details of models used in our evaluation. Context refers to the maximum context window size in tokens. Underlined models are closed source.}
    \label{tab:model_details}
    \begin{tabular}{lccc}
    \toprule
    \textbf{Model} & \textbf{Size (B)} & \textbf{Context} \\
    \midrule
    \underline{Gemini 2.5 Pro} & - & 1M \\
    \underline{GPT-4o} & - & 128K \\
    \underline{Claude 3.5 Sonnet} & - & 200K \\
    DeepSeek V3 & 685 & 164K \\
    Qwen 2.5 & 72 & 33K \\
    Llama 4 Maverick & 400 & 1M \\
    Gemma 3 & 27 & 131K \\
    \bottomrule
    \end{tabular}
\end{table}

\section{Evaluation}
We evaluate on 7 state-of-the-art models and 2 baselines. For the baselines, we evaluate a model based on a random choice of options and a model based on keyword extraction. For the other models, we use the OpenRouter\footnote{\url{https://openrouter.ai}} API to run all the experiments. The temperature is set to 0.0. We use the validation set to tune the system prompt. We repeat each inference a maximum of 6 times until there is no error. We evaluate a diverse set of proprietary and open-source LLMs—ranging from around 30 to over 600 billion parameters—to assess their capabilities on this task, as detailed in Table \ref{tab:model_details}. 

\subsection{Categories}
We evaluated all models in the different language categories in Table \ref{tab:main}. Generally speaking, Gemini 2.5 Pro achieves the highest average score. Across different categories, Gemini 2.5 Pro also achieves the highest score in 5 out of the 6 categories. Smaller LLMs like Gemma 3, with only 27B parameters, achieves decent results on the benchmark. We note that Claude Sonnet 3.5 results in many errors, which caused its score to be lower for some categories. 

\begin{figure}[!htp]
    \centering
    \includegraphics[width=\linewidth]{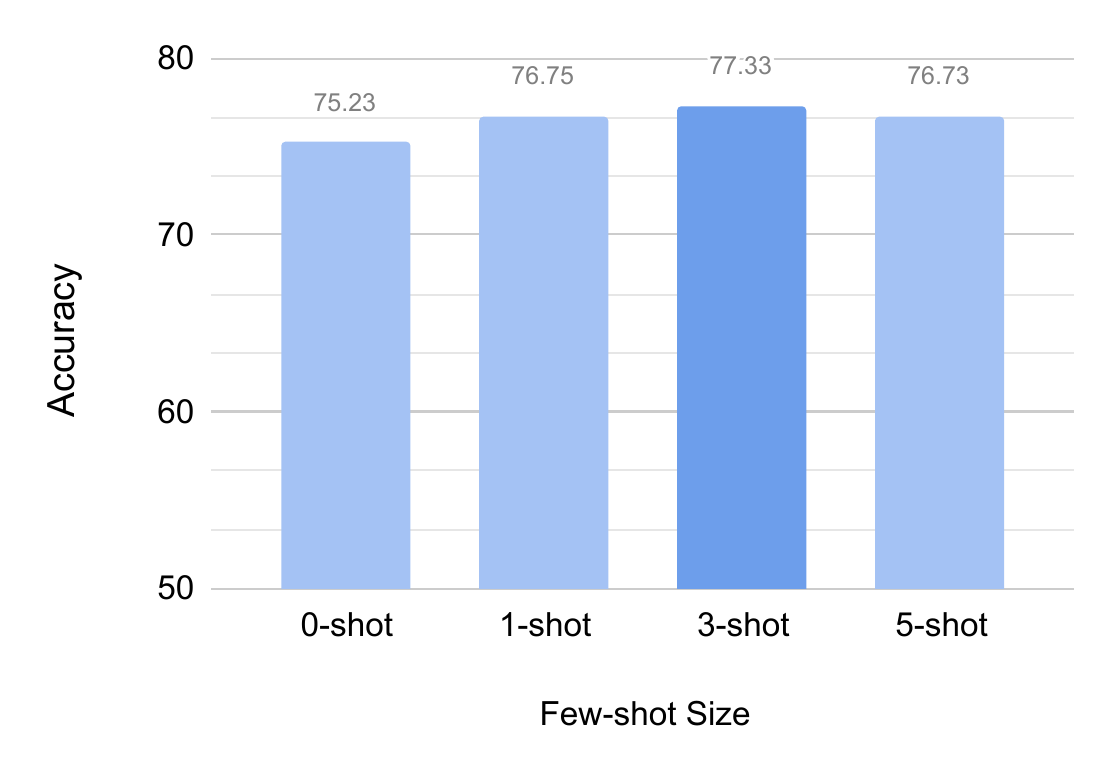}
    \caption{Few-shot results with 0, 1, 3, and 5 -shot examples using the Gemini 2.5 Pro model. }
    \label{fig:fewshot}
\end{figure}

\begin{figure*}[t!]
    \centering
    \includegraphics[width=0.8\linewidth]{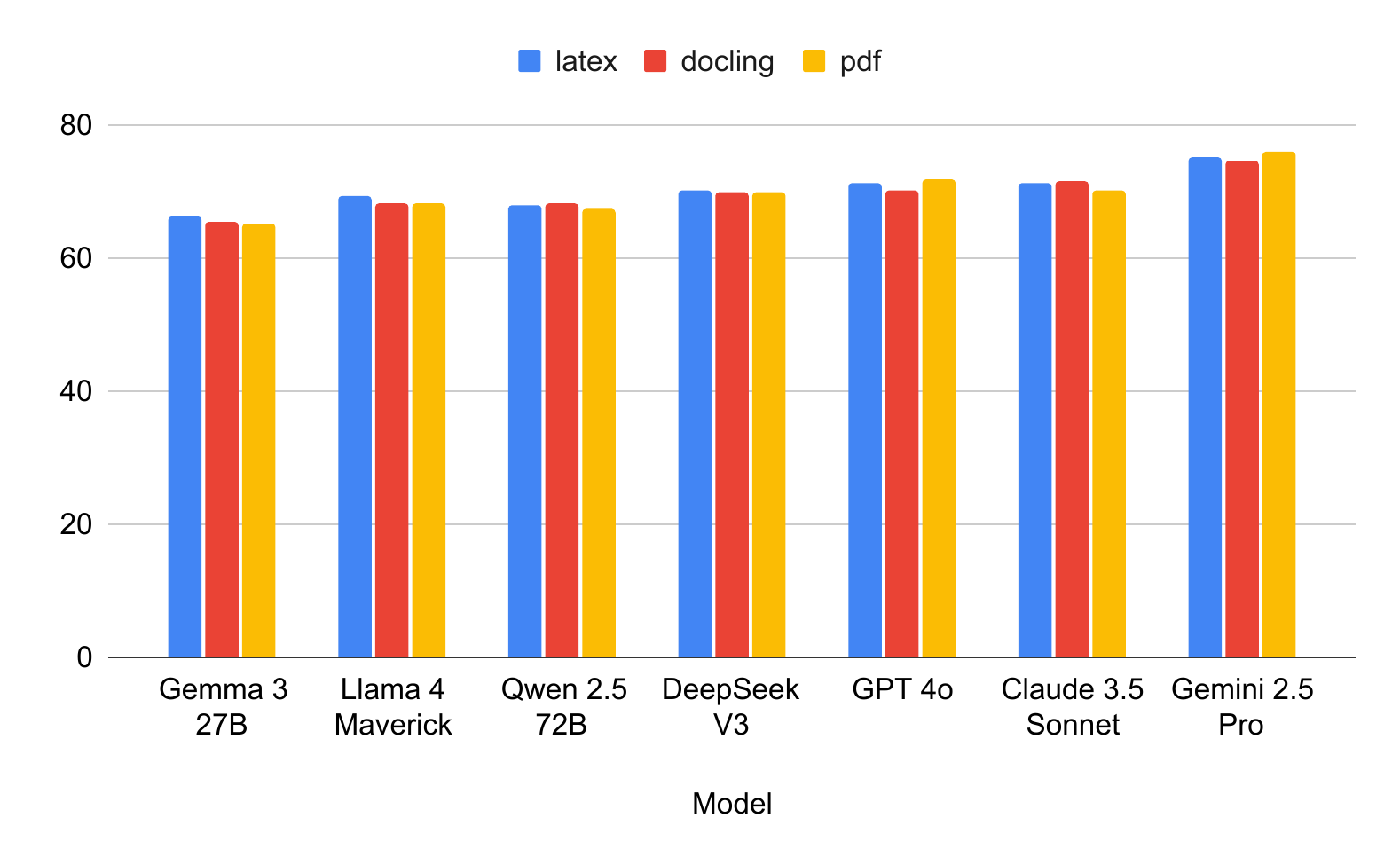}
    \caption{Latex vs. PDF vs. Docling input formats results across all models. }
    \label{fig:pdf}
\end{figure*}

\subsection{Few-shot}
Since our benchmarks requires structured formatting, we test with different n-shot examples. Since processing multiple papers in a few-shot is expensive, we rely on synthetic examples creation and only evaluate the results using our top model, which is Gemini 2.5 Pro. We create the examples using a template and fill the attributes randomly (see Appendix \ref{app:synthetic} for more details). In Figure \ref{fig:fewshot}, we show that providing examples improve the results compared to zero-shot. In particular, 3-shot examples achieve the highest gain in results compared to zero-shot. 

\subsection{Input format}
We experiment with three approaches for test input, using LaTeX, PDF text, and structured output using Docling \cite{docling}. We are interested in validating the performance on other input formats, as, in many scenarios, the LaTeX source may not necessarily be available. To extract the text content of a PDF, we use Python pdfplumber\footnote{\url{https://pypi.org/project/pdfplumber/}}. We observe that for smaller models, the LaTeX format is slightly better compared to PDF and Docling. However, we don't observe a clear trend across all models. 

\begin{figure}[t]
    \centering
    \includegraphics[width=\linewidth]{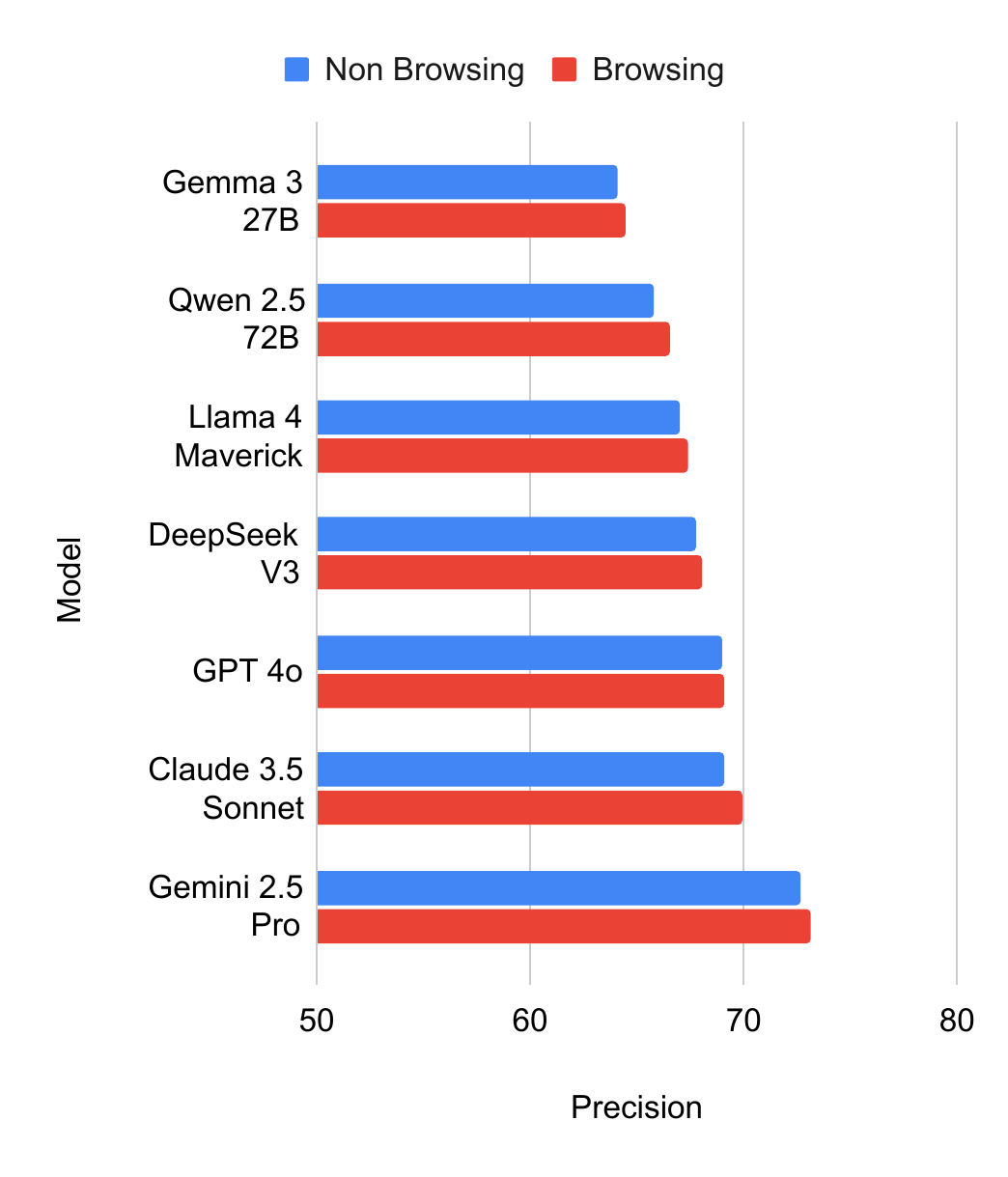}
    \caption{Browsing vs. no Browsing for all models in our evaluation benchmarks. }
    \label{fig:browsing}
\end{figure}%

\subsection{Browsing}
Some annotated attributes might not exist in the papers. For example, the license attribute is mostly extracted from the repository where the dataset is hosted. To allow all the models to browse, we use the extracted metadata from the non-browsing approach and the page where the dataset is hosted to predict the updated metadata attributes. For repositories that contain a README.md file, like GitHub and HuggingFace, we fetch the file directly from the repository. In Figure \ref{fig:browsing}, we show the results across all the models. We observe a clear improvement when using browsing for all of the models. We note that the precision highlights more the effect of browsing, as it indicates whether models can reliably predict attributes that are not in the paper. Our experiences show that recall is affected by browsing, as the attributes that exist in the paper might get deflected by browsing data. 

\begin{table}[!htp]
\centering
\small
\caption{Model scores across three different constraints for the length of answer output (Low, Mid, High). The values are normalized by all fields in the metadata.}
\label{tab:length}
\begin{tabular}{lccc}
\toprule
\textbf{Model}             & \textbf{Low}                 & \textbf{Mid}                 & \textbf{High}                \\
\midrule
\textbf{Gemma 3 27B}       & \cellcolor[HTML]{70C59B}0.99 & \cellcolor[HTML]{B8E2CD}0.96 & \cellcolor[HTML]{FFFFFF}0.93 \\
\textbf{Qwen 2.5 72B}      & \cellcolor[HTML]{57BB8A}1.00 & \cellcolor[HTML]{88CFAC}0.98 & \cellcolor[HTML]{E8F6EF}0.94 \\
\textbf{DeepSeek V3}       & \cellcolor[HTML]{57BB8A}1.00 & \cellcolor[HTML]{70C59B}0.99 & \cellcolor[HTML]{D0ECDE}0.95 \\
\textbf{GPT 4o}            & \cellcolor[HTML]{70C59B}0.99 & \cellcolor[HTML]{B8E2CD}0.96 & \cellcolor[HTML]{D0ECDE}0.95 \\
\textbf{Claude 3.5 Sonnet} & \cellcolor[HTML]{70C59B}0.99 & \cellcolor[HTML]{70C59B}0.99 & \cellcolor[HTML]{B8E2CD}0.96 \\
\textbf{Llama 4 Maverick}  & \cellcolor[HTML]{57BB8A}1.00 & \cellcolor[HTML]{88CFAC}0.98 & \cellcolor[HTML]{B8E2CD}0.96 \\
\textbf{Gemini 2.5 Pro}    & \cellcolor[HTML]{57BB8A}1.00 & \cellcolor[HTML]{88CFAC}0.98 & \cellcolor[HTML]{A0D9BD}0.97
\\
\bottomrule
\end{tabular}
\end{table}

\begin{table*}[!htp]
\centering
\small 
\caption{Model comparison across different validation groups using F1 score. }
\begin{tabular}{lccccc}
\toprule
\textbf{Model} & \textbf{Diversity} & \textbf{Accessibility} & \textbf{Content} & \textbf{Evaluation} & \textbf{Average} \\
\midrule
\textbf{Random} & 46.83 & 24.95 & 27.64 & 34.32 & 33.43 \\
\textbf{Keyword} & 67.20 & 34.78 & 44.43 & 49.26 & 48.91 \\
\textbf{Qwen 2.5 72B} & 79.50 & 67.00 & 68.85 & 58.73 & 68.52 \\
\textbf{Gemma 3 27B} & 81.08 & 66.00 & 62.14 & 67.46 & 69.17 \\
\textbf{Llama 4 Maverick} & 81.08 & \underline{68.04} & 70.88 & 62.28 & 70.57 \\
\textbf{DeepSeek V3} & 82.54 & 67.55 & 70.99 & 63.60 & 71.17 \\
\textbf{Claude 3.5 Sonnet} & 78.57 & 67.35 & \underline{72.33} & \underline{70.48} & 72.18 \\
\textbf{GPT 4o} & \underline{84.13} & 67.78 & 70.64 & \textbf{71.69} & \underline{73.56} \\
\textbf{Gemini 2.5 Pro} & \textbf{87.17} & \textbf{69.24} & \textbf{78.75} & 70.42 & \textbf{76.40} \\
\bottomrule
\end{tabular}
\label{tab:model_comparison}
\end{table*}

\subsection{Length Enforcing}
We use the min and max of the answer to check if the answer from a given LLM respects the required length from the schema. Additionally, to see the effect of increasing the length constraints, we define the following three granularities:

\begin{enumerate}
    \item \textbf{Low} this is the standard type of constraint used in all previous experiments. It is considered more relaxed compared to others. 
    \item \textbf{Mid}  a medium constraint used to decrease the range of the following attributes: Name, Description, Provider, Derived From, and Tasks.
    \item \textbf{High} similar to Mid, we use the same attributes but with more stricter range. 
\end{enumerate}

As an example, the attribute \textit{Description}, will have the \textit{answer\_max} as (50, 25, 12) for low, mid, and high, respectively. In Table \ref{tab:length}, we highlight the results across all the models for low, mid, and high length constraints. Still, Gemini 2.5 Pro achieves the highest adherence to high constraints. We note that LLMs that cause many errors in structured output will adhere more to constraints as the output schema will be empty.  

\subsection{Context Length}
\begin{figure}
    \centering
    \includegraphics[width=\linewidth]{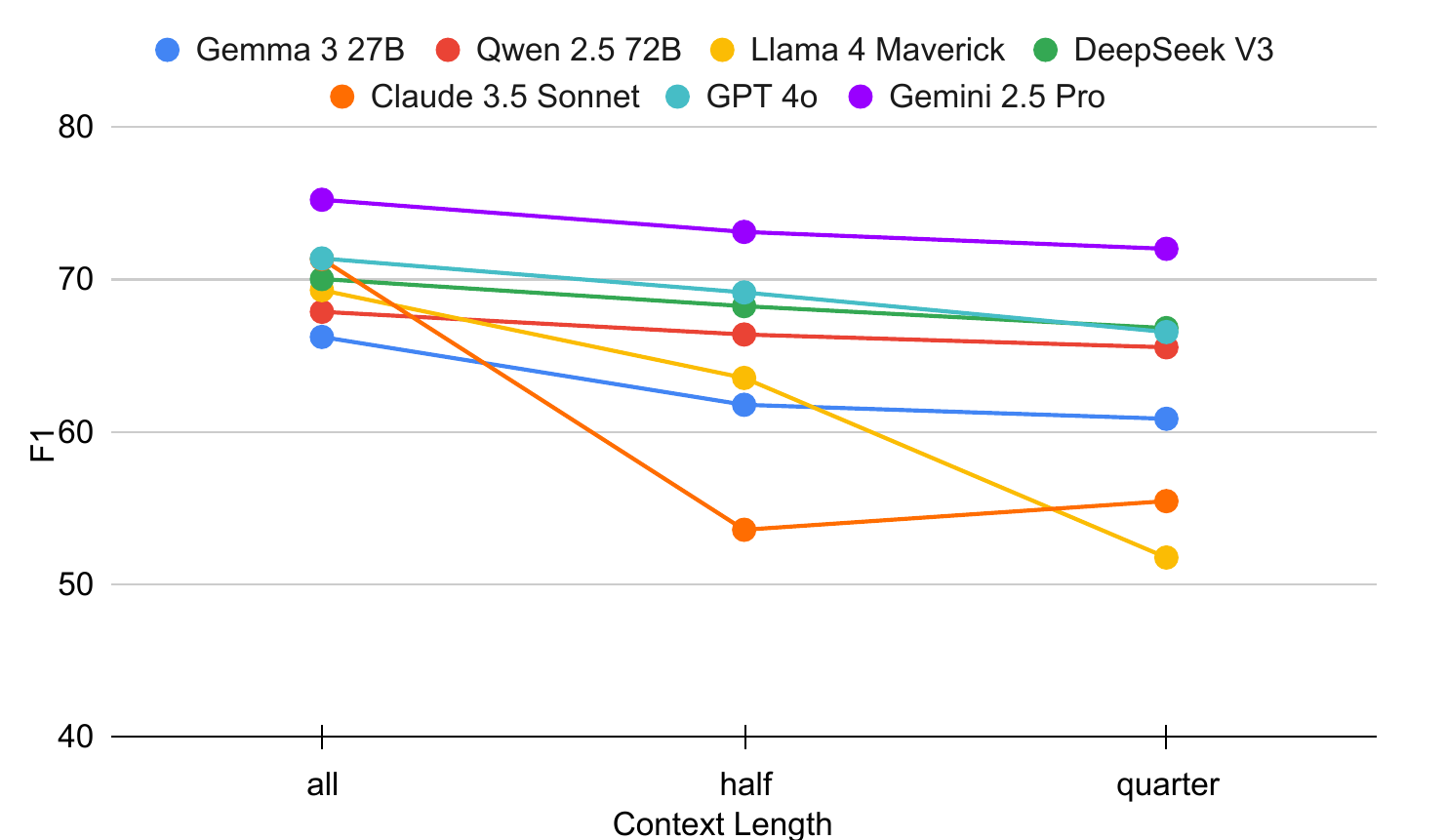}
    \caption{The effect of changing the context length on the results of all models. }
    \label{fig:context}
\end{figure}

In Figure \ref{fig:context}, we show the effect of varying the context length on the results of all models. Interestingly, Gemini 2.5 Pro can still achieve competitive results by only selecting half or a quarter of the context. This shows that most of the metadata can be extracted at the upper part of the paper. A similar trend can also be seen for GPT-4o, Qwen, and DeepSeek. For other models, the results are affected significantly, especially for Llama and Claude Sonnet, where the results decrease dramatically. Also, we noticed the error frequency increased for such models when using a smaller context.

\subsection{Validation Groups}
\begin{figure}[!htp]
    \centering
    \includegraphics[width=0.9\linewidth]{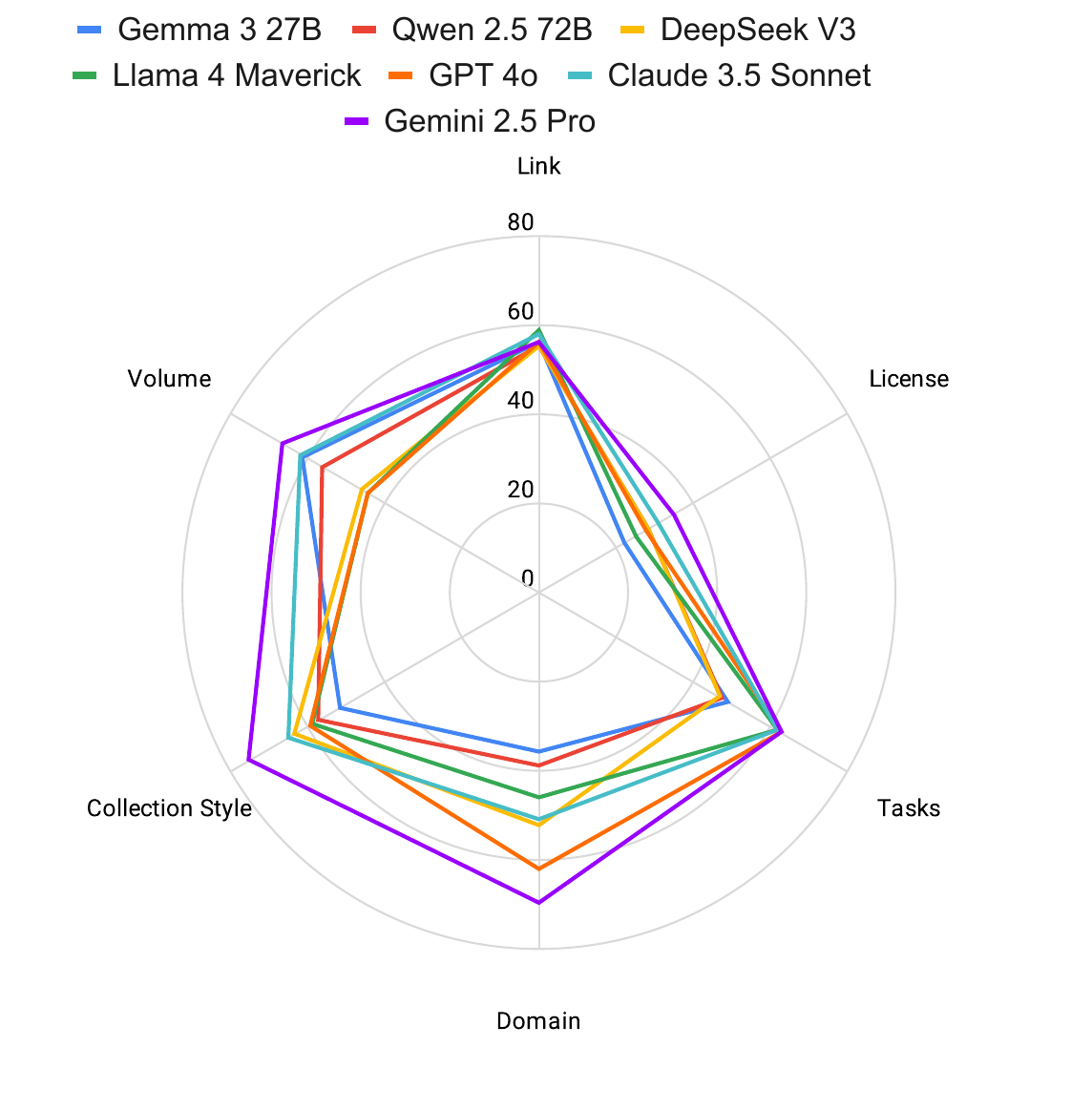}
    \caption{Precision across 6 different metadata attributes (Link, Volume, License, Collection Style, Domain, and Tasks). }
    \label{fig:radar}
\end{figure}

Our metadata attributes are divided into four validation groups, which are diversity, accessibility, content, and evaluation (see Appendix \ref{app:groups} and Figure \ref{fig:groups}). In Table \ref{tab:model_comparison}, we compare the results of different models on the different validation groups. Gemini 2.5 Pro achieves the highest results in 3 out of the 4 groups. In general, we see all models struggle to achieve high scores for accessibility, which requires extracting the link, license, and host, etc. On the other hand, diversity is the easiest group to extract attributes for. 
Figure \ref{fig:radar} shows the results of 6 attributes across different models. While Gemini 2.5 Pro achieves the highest average score, it doesn't achieve the highest scores across all attributes. Interestingly, some LLMs might achieve the highest scores on a single attribute like Llama on the \textit{Link} attribute. 

\section{Related Work}

The exponential growth of research data has made metadata extraction increasingly critical \cite{gebru2021datasheets, mahadevkar2024exploring, yang2025impact}. We examine the evolution and current state of metadata extraction research across three areas.

\paragraph{Evolution of Metadata Extraction Approaches}
Early systems relied on rule-based and traditional machine learning methods. CERMINE \cite{tkaczyk2015cermine} employed modular architectures for bibliographic extraction, building on methods like PDFX \cite{constantin2013pdfx}. FLAG-PDFe \cite{ahmad2020flag} introduced feature-oriented frameworks using SVMs for scientific publications.

Deep learning marked a paradigm shift in this field. \cite{an2017citation} introduced neural sequence labeling for citation metadata extraction. Multimodal approaches \cite{boukhers2022vision,liu2018automatic} integrated NLP with computer vision to handle layout diversity in PDF documents. Recent work includes domain-specific applications in chemistry \cite{schilling2024text,zhu2022pdfdataextractor}, HPC \cite{schembera2021like}, and cybersecurity \cite{pizzolante2024unlocking}, with some exploring LLMs for metadata extraction. Cross-lingual approaches have addressed language-specific challenges, including Korean complexities \cite{kong2022annotated}, Persian \cite{rahnama2020automatic}, and Arabic NLP resource cataloging through Masader \cite{alyafeai2021masader, altaher2022masader}.

\paragraph{Standardization Efforts}
\cite{gebru2021datasheets} proposed standardized templates for ML dataset documentation, influencing practices across digital heritage \cite{alkemade2023datasheets}, healthcare \cite{rostamzadeh2022healthsheet}, energy \cite{heintz2023datasheets}, art \cite{srinivasan2021artsheets}, and earth science \cite{connolly2025datasheets}. DescribeML \cite{giner2022describeml} provided a domain-specific language with IDE integration for practical implementation\footnote{\url{https://code.visualstudio.com/}}.

\paragraph{Evaluation Benchmarks}
Several benchmarks exist for metadata extraction evaluation. PARDA \cite{fan2019parda} provides annotated samples across domains and formats. The unarXive corpus \cite{saier2020unarxive} represents one of the largest scholarly datasets with full-text publications and metadata links. DocBank \cite{li2020docbank} and \cite{meuschke2023benchmark} offer additional evaluation frameworks. However, these focus on general paper attributes (title, authors, abstract) rather than detailed dataset characteristics like volume, license, and subsets that our work addresses.

\section{Conclusion}

This paper introduced a methodology for using LLMs to extract and validate metadata from scientific papers. Through our framework, which extracts around 30 distinct metadata attributes, we offer a more robust approach to automating metadata extraction. We experiment with multiple approaches, including different input formats, few-shot examples, and browsing. We also highlight the effect of length constraints and how LLMs still struggle to follow strict instructions. Throughout our experiments, we show recent advancements in processing long context, especially in flagship models like Gemini 2.5 Pro and GPT-4o that continue to achieve better results. We also release a benchmark of 126 papers manually annotated to facilitate research in metadata extraction. This work contributes not only to the advancement of metadata extraction techniques but also to the broader goal of making scientific research more transparent, accessible, and reusable. 

\section*{Limitations}

In this section, we provide some possible limitations and our proposed mitigation and possible future suggestions to improve our work:

\begin{enumerate}
    \item \textbf{Cost} Processing thousands of tokens for a given paper increases the cost. Interestingly, We showed in this paper that most metadata can be extracted using a smaller context. One possible future direction is to make this process more cost-effective by reducing the context size. The context size can be reduced by using some kind of early skimming using a lighter/less expensive LLM.  

    \item \textbf{Length Enforcing} Length constraining is a difficult problem, and current LLMs are not capable of reliably predicting the exact number of tokens \cite{muennighoff2025s1}. As a possible direction, we can use precise controlling methods like \cite{butcher2025precise} to extract structured data with better adherence to length constraints. If LLMs become optimal at length enforcing, they can become more cost-effective as we can generate as many tokens as we need. 

    \item \textbf{Source Code Availability} Our approach mostly depends on the availability of LaTeX source code. To mitigate this, we also compare the results for other input formats like PDF and structured format using Docling. However, using such an approach might be difficult to scale, especially due to time constraints. As a future direction, we will explore how to improve the skimming process and cleaning of PDF content. We can utilize this method to scale our approach to thousands of papers. 
    
\end{enumerate}

\section*{Acknowledgments}
The research reported in this publication was supported by funding from King Abdullah University of Science and Technology (KAUST) - Center of Excellence for Generative AI, under award number 5940.

We want also to thank Ali Fadel and Amr Keleg for the useful discussions. 
\bibliography{custom}

\appendix
\section{Datasets}

In Total, we annotated 132 papers including validation and testing. Here is a list of the paper datasets: RuBQ \cite{RuBQ}, RuCoLA \cite{RuCoLA}, Slovo \cite{Slovo}, WikiOmnia \cite{WikiOmnia}, DaNetQA \cite{DaNetQA}, RuSemShift \cite{RuSemShift}, REPA \cite{REPA}, SberQuAD \cite{SberQuAD}, RuBLiMP \cite{RuBLiMP}, Golos \cite{Golos}, RFSD \cite{RFSD}, RuDSI \cite{RuDSI}, HeadlineCause \cite{HeadlineCause}, RusTitW \cite{RusTitW}, CRAFT \cite{CRAFT}, Gazeta \cite{Gazeta}, RuBia \cite{RuBia}, RusCode \cite{RusCode}, Russian Jeopardy \cite{Russian_Jeopardy}, Russian Multimodal Summarization \cite{Russian_Multimodal_Summarization}, NEREL \cite{NEREL}, HISTOIRESMORALES \cite{HISTOIRESMORALES}, PxSLU \cite{PxSLU}, French COVID19 Lockdown Twitter Dataset \cite{French_COVID19_Lockdown_Twitter_Dataset}, MEDIA with Intents \cite{MEDIA_with_Intents}, FrenchMedMCQA \cite{FrenchMedMCQA}, FRASIMED \cite{FRASIMED}, FQuAD1.1 \cite{FQuAD1.1}, FIJO \cite{FIJO}, BSARD \cite{BSARD}, FairTranslate \cite{FairTranslate}, 20min-XD \cite{20min-XD}, Alloprof \cite{Alloprof}, FREDSum \cite{FREDSum}, Vibravox \cite{Vibravox}, MTNT \cite{MTNT}, PIAF \cite{PIAF}, FrenchToxicityPrompts \cite{FrenchToxicityPrompts}, OBSINFOX \cite{OBSINFOX}, CFDD \cite{CFDD}, FREEMmax \cite{FREEMmax}, FQuAD2.0 \cite{FQuAD2.0}, HellaSwag \cite{HellaSwag}, GPQA \cite{GPQA}, GoEmotions \cite{GoEmotions}, SQuAD 2.0 \cite{SQuAD_2.0}, LAMBADA \cite{LAMBADA}, ClimbMix \cite{ClimbMix}, RACE \cite{RACE}, MMLU-Pro \cite{MMLU-Pro}, BoolQ \cite{BoolQ}, GSM8K \cite{GSM8K}, HotpotQA \cite{HotpotQA}, SQuAD \cite{SQuAD}, RefinedWeb \cite{RefinedWeb}, MMLU \cite{MMLU}, PIQA \cite{PIQA}, BRIGHT \cite{BRIGHT}, HLE \cite{HLE}, TinyStories \cite{TinyStories}, WinoGrande \cite{WinoGrande}, SciQ \cite{SciQ}, TriviaQA \cite{TriviaQA}, JEC \cite{JEC}, JParaCrawl \cite{JParaCrawl}, KaoKore \cite{KaoKore}, llm-japanese-dataset v0 \cite{llm-japanese-dataset_v0}, JaLeCoN \cite{JaLeCoN}, JaQuAD \cite{JaQuAD}, JAFFE \cite{JAFFE}, JaFIn \cite{JaFIn}, Japanese Fake News Dataset \cite{Japanese_Fake_News_Dataset}, JMultiWOZ \cite{JMultiWOZ}, Japanese Web Corpus \cite{Japanese_Web_Corpus}, J-CRe3 \cite{J-CRe3}, JSUT \cite{JSUT}, Japanese Word Similarity Dataset \cite{Japanese_Word_Similarity_Dataset}, STAIR Captions \cite{STAIR_Captions}, Arukikata Travelogue \cite{Arukikata_Travelogue}, JTubeSpeech \cite{JTubeSpeech}, JCoLA \cite{JCoLA}, JESC \cite{JESC}, JDocQA \cite{JDocQA}, Jamp \cite{Jamp}, 101 Billion Arabic Words Dataset \cite{101_Billion_Arabic_Words_Dataset}, WinoMT \cite{WinoMT}, ArabicMMLU \cite{ArabicMMLU}, CIDAR \cite{CIDAR}, Belebele \cite{Belebele}, MGB-2 \cite{MGB-2}, ANETAC \cite{ANETAC}, TUNIZI \cite{TUNIZI}, Shamela \cite{Shamela}, POLYGLOT-NER \cite{POLYGLOT-NER}, DODa \cite{DODa}, LASER \cite{LASER}, MGB-3 \cite{MGB-3}, Arap-Tweet \cite{Arap-Tweet}, FLORES-101 \cite{FLORES-101}, Transliteration \cite{Transliteration}, ADI-5 \cite{ADI-5}, Maknuune \cite{Maknuune}, EmojisAnchors \cite{EmojisAnchors}, Calliar \cite{Calliar}, LABR \cite{LABR}, ACVA \cite{ACVA}, ATHAR \cite{ATHAR}, OpenITI-proc \cite{OpenITI-proc}, AraDangspeech \cite{AraDangspeech}, Arabic-Hebrew TED Talks Parallel Corpus \cite{Arabic-Hebrew_TED_Talks_Parallel_Corpus}, ARASPIDER \cite{ARASPIDER}, XNLI \cite{XNLI}, X-stance \cite{X-stance}, DiS-ReX \cite{DiS-ReX}, RELX \cite{RELX}, MultiSubs \cite{MultiSubs}, MEE \cite{MEE}, XCOPA \cite{XCOPA}, MLQA \cite{MLQA}, M2DS \cite{M2DS}, XOR-TyDi \cite{XOR-TyDi}, Multilingual Hate Speech Detection Dataset \cite{Multilingual_Hate_Speech_Detection_Dataset}, MINION \cite{MINION}, SEAHORSE \cite{SEAHORSE}, Mintaka \cite{Mintaka}, Multi2WOZ \cite{Multi2WOZ}, MTOP \cite{MTOP}, X-RiSAWOZ \cite{X-RiSAWOZ}, PRESTO \cite{PRESTO}, LAHM \cite{LAHM}, MARC \cite{MARC}, and MLSUM \cite{MLSUM}.

\section{Costs}
In Table \ref{tab:costs}, we show the costs for each model in terms of input tokens, output tokens, and cost in one single experiment. In general, we observe that Claude is more costly while processing the least number of tokens. Gemini 2.5 Pro achieves the best cost vs performance results as it is cheaper than other models but achieves better results across different experiments. In total, it costs around 20 USD to run the evaluation and extract the metadata for all the papers. 
\section{Errors}
\begin{figure}
    \centering
    \includegraphics[width=\linewidth]{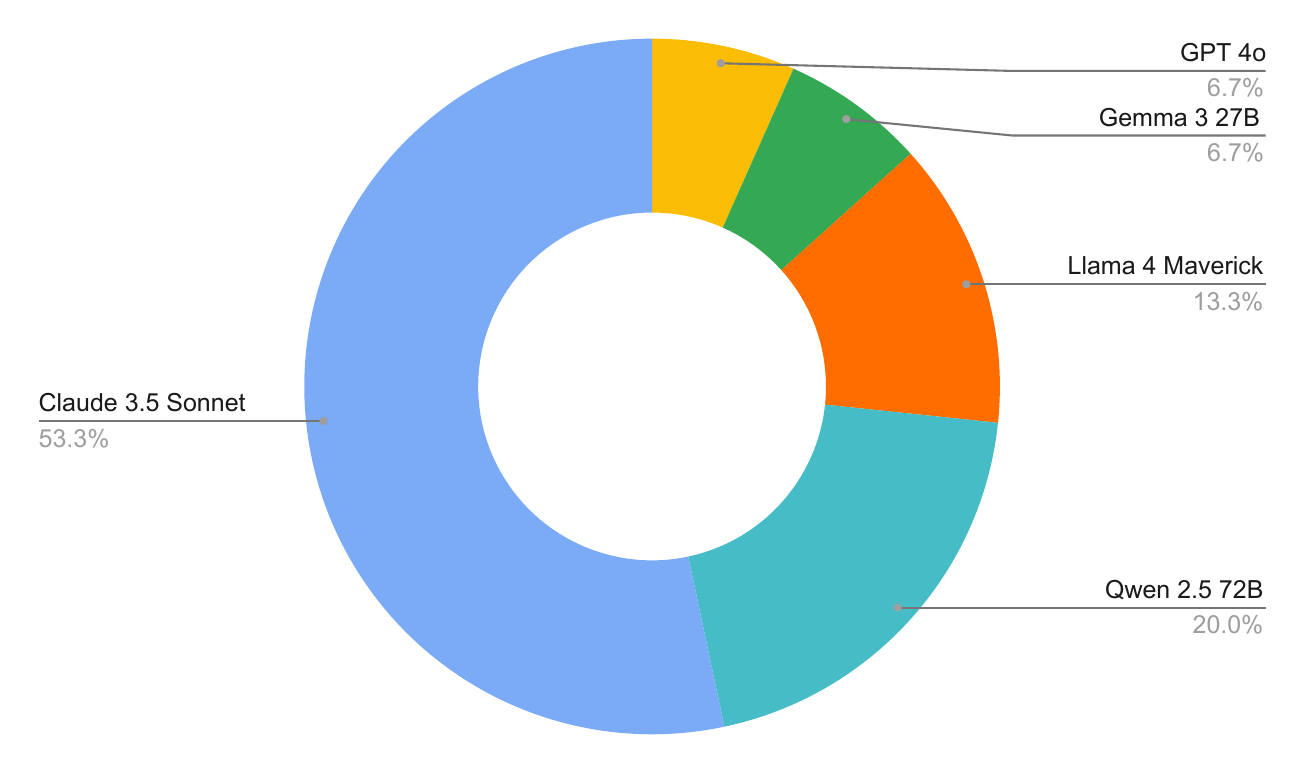}
    \caption{Distribution of the number of errors per model for one experiment. This graph only shows the models with at least one error. }
    \label{fig:errors}
\end{figure}

In Figure \ref{fig:errors}, we show the number of errors for each model in a single experiment. The major proportion of errors comes from Claude Sonnet 3.5. In general, we run more than 1K inference requests to the OpenRouter API, and around 16 errors happened. Most of the errors occur due to failing to read the generated JSON file. 

\section{Input Format Processing Time}
\begin{figure}[!htp]
    \centering
    \includegraphics[width=\linewidth]{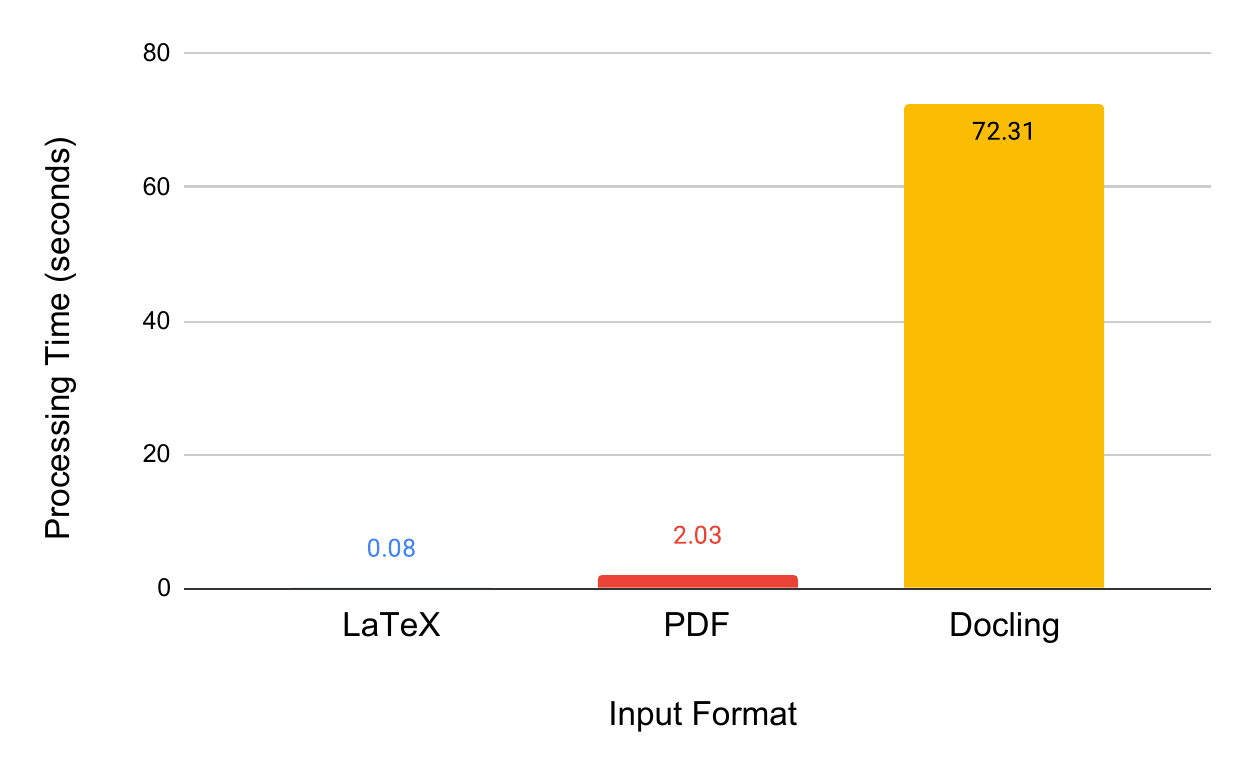}
    \caption{Average processing time comparison between different input formats. LaTeX source processing (0.08s) is most efficient, followed by pdfplumber-based PDF extraction (2.03s), while Docling structured parsing (72.31s) requires substantially more computation time.}
    \label{fig:process_time}
\end{figure}

We analyze the processing time of each input format. Figure~\ref{fig:process_time} shows the average of processing time where we applied each of these methods on the subset of the testset. The figure shows the varying differences in preprocessing time across these methods. LaTeX source processing is remarkably efficient (0.08s), while PDF extraction via pdfplumber is reasonably fast (2.03s). In contrast, Docling processing requires significantly more compute time (72.31s), highlighting important efficiency trade-offs when scaling metadata extraction to large document collections.

\section{Model's Access}
In Table \ref{tab:models}, we highlight the models used for evaluation. We use a collection of closed and open models. We use the OpenRouter API to perform inference on all models. 
\begin{table*}[]
    \centering
    \small
    \caption{Models used for evaluation and their respective links in OpenRouter.}
    \label{tab:models}
    \begin{tabular}{ll}
    \toprule
    \textbf{Model} & \textbf{Link} \\ 
    \midrule
        GPT 4o & \url{https://openrouter.ai/openai/gpt-4o} \\
        Claude 3.5 Sonnet & \url{https://openrouter.ai/anthropic/claude-3.5-sonnet} \\
        Gemini 2.5 Pro & \url{https://openrouter.ai/google/gemini-2.5-pro-preview-03-25} \\
        DeepSeek V3 & \url{https://openrouter.ai/deepseek/deepseek-chat-v3-0324} \\
        Llama 4 Maverick & \url{https://openrouter.ai/meta-llama/llama-4-maverick} \\
        Gemma 3 27B & \url{https://openrouter.ai/google/gemma-3-27b-it} \\
        Qwen2.5 72B & \url{https://openrouter.ai/qwen/qwen-2.5-72b-instruct} \\
        \bottomrule
    \end{tabular}
\end{table*}
\label{sec:appendix}

\begin{table*}[]
\centering
\small
\caption{Input, output, and total tokens and cost in USD for running one experiment on all test sets using LaTeX as input. We use OpenRouter to estimate the number of input and output tokens.}
\label{tab:costs}
\begin{tabular}{lcccc}
\toprule
\textbf{Model} & \textbf{Input Tokens} & \textbf{Output Tokens} & \textbf{Total Tokens} & \textbf{Cost (USD)} \\
\midrule
\textbf{Gemma 3 27B} & 2162756 & {85738} & {2248494} & 0.23 \\
\textbf{Qwen 2.5 72B} & 2163466 & 77643 & 2241109 & 0.29 \\
\textbf{Llama 4 Maverick} & 2159318 & 71856 & 2231174 & 0.41 \\
\textbf{DeepSeek V3} & 2163754 & 77280 & 2241034 & 0.72 \\
\textbf{Gemini 2.5 Pro} & {2163823} & {160811} & {2324634} & 4.31 \\
\textbf{GPT 4o} & {2163823} & 80514 & 2244337 & {6.21} \\
\textbf{Claude 3.5 Sonnet} & {2163934} & 74944 & 2238878 & \textbf{7.62} \\
\bottomrule
\end{tabular}

\end{table*}

\section{Synthetic Template Generation}
\label{app:synthetic}
For few-shot experiments, we create synthetic examples, which are short paper templates that can be filled using the different attributes in each metadata schema. In Figure \ref{fig:template}, we show an example of a template.  The language table is used for the multilingual schema, where we create a table for the language subsets of the datasets. To add variance in the results, we sample different options for attributes like unit, task, collection style, and domain. We also sample a different number each time for the volume. For each schema category in our dataset we generate five examples. 

\begin{figure}
    \centering
    \small
    \begin{tcolorbox}[colback=black!10, boxrule=0pt]
\{name\}: A \{task\} dataset for \{schema\}\newline
\{authors\} \newline
\{affs\} \newline
\{name\}, is a \{schema\} \{task\} dataset, that contains \{volume\} \{unit\}.
\{language\_table\}
\{provider\_stmt\} The dataset was collected from \{collection\_style\} of \{domain\} in \{year\}. 
The dataset is publicly available through this link \{link\}. \{license\_stmt\}.
\{hf\_stmt\}.
\end{tcolorbox}
\caption{Template-based few-shot example creation.}
\label{fig:template}
\end{figure}

\section{Validation Groups}
\label{app:groups}
\begin{figure*}
    \centering
    \includegraphics[width=0.9\linewidth]{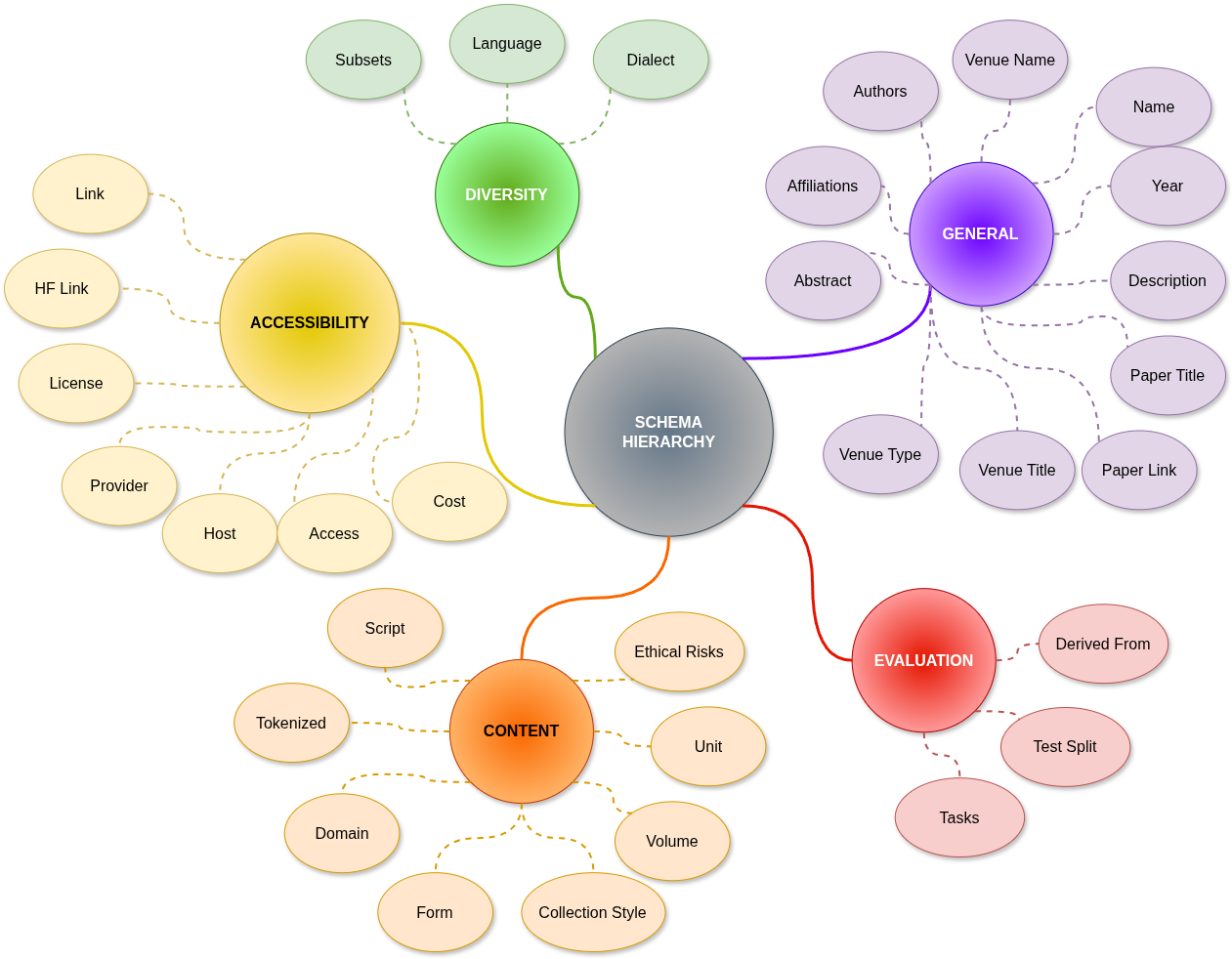}
    
    \caption{Schema validation groups and their associated attributes for the Arabic metadata. 
    }
    \label{fig:groups}
\end{figure*}

We have four groups that gather similar metadata attributes. Figure \ref{fig:groups} shows the different validation groups in the Arabic subset of the dataset. The general group is not used for validation, as it shows easily extractable attributes. Each group covers attributes that are similar in terms of the grouping function. 

\begin{table*}[]
\centering
\small
\caption{Results across different attributes that exist in all categories. For better visualization, we remove the versions and naming conventions of each model.}
\label{tab:individual}
\begin{tabular}{lccccccc}
\toprule
\textbf{Model} & \textbf{Llama} & \textbf{Gemini} & \textbf{Qwen} & \textbf{DeepSeek} & \textbf{GPT} & \textbf{Gemma} & \textbf{Claude} \\
\midrule
\textbf{Link} & 58.73 & 56.35 & 55.56 & 55.56 & 56.35 & 56.35 & 57.94 \\
\textbf{HF Link} & 47.62 & 47.62 & 44.44 & 47.62 & 48.41 & 45.24 & 48.41 \\
\textbf{License} & 25.40 & 34.92 & 28.57 & 28.57 & 27.78 & 22.22 & 30.95 \\
\textbf{Language} & 89.68 & 93.65 & 88.10 & 90.48 & 92.86 & 88.89 & 84.13 \\
\textbf{Domain} & 46.03 & 69.84 & 38.89 & 52.38 & 61.90 & 35.71 & 50.79 \\
\textbf{Form} & 94.44 & {97.62} & 95.24 & {96.03} & 96.03 & 89.68 & 88.89 \\
\textbf{Collection Style} & 58.73 & 75.40 & 57.14 & 63.49 & 59.52 & 51.59 & 65.08 \\
\textbf{Volume} & 44.44 & 66.67 & 56.35 & 46.03 & 44.44 & 61.11 & 61.90 \\
\textbf{Unit} & 69.05 & 70.63 & 53.17 & 59.52 & 54.76 & 58.73 & 59.52 \\
\textbf{Ethical Risks} & 83.33 & 80.16 & 82.54 & 81.75 & 80.16 & 34.13 & 84.13 \\
\textbf{Provider} & 50.00 & 53.17 & 50.00 & 52.38 & 50.00 & 48.41 & 47.62 \\
\textbf{Derived From} & 53.17 & 61.90 & 53.97 & 66.67 & 66.67 & 73.81 & 65.08 \\
\textbf{Tokenized} & 92.86 & 85.71 & 91.27 & 92.86 & 91.27 & {95.24} & {90.48} \\
\textbf{Host} & 68.25 & 69.84 & 68.25 & 69.05 & 67.46 & 65.87 & 69.84 \\
\textbf{Access} & {97.62} & 92.06 & {97.62} & {96.03} & {96.83} & 94.44 & 89.68 \\
\textbf{Cost} & {96.83} & {100.00} & {100.00} & {96.83} & {100.00} & {99.21} & {100.00} \\
\textbf{Test Split} & 71.43 & 86.51 & 74.60 & 76.98 & 85.71 & 79.37 & 84.13 \\
\textbf{Tasks} & 61.90 & 62.70 & 47.62 & 46.83 & 62.70 & 49.21 & 61.90 \\
\bottomrule
\end{tabular}

\end{table*}

\begin{table*}[]
\centering
\small
\caption{Difference between Browsing and non-browsing models in all attributes. Model names are shortened for better visualization.}
\label{tab:individual-diff}
\begin{tabular}{lccccccc}
\toprule
\textbf{Model} & \textbf{Llama} & \textbf{Gemini} & \textbf{Qwen} & \textbf{DeepSeek} & \textbf{GPT} & \textbf{Gemma} & \textbf{Claude} \\
\midrule
\textbf{Link} & -0.79 & -4.76 & -5.56 & {0.79} & -6.35 & -1.59 & -2.38 \\
\textbf{HF Link} & 0.79 & 0.79 & {1.59} & 0.00 & -0.79 & -0.79 & 0.00 \\
\textbf{License} & {15.08} & {15.08} & {19.05} & {7.14} & {15.08} & {13.49} & {21.43} \\
\textbf{Language} & -0.79 & 0.79 & 0.00 & 0.00 & 0.00 & -0.79 & 0.79 \\
\textbf{Domain} & 0.79 & 0.79 & 0.00 & 0.00 & 0.00 & 1.59 & {1.59} \\
\textbf{Form} & 0.00 & 0.00 & 0.79 & 0.00 & 0.00 & 0.00 & 0.79 \\
\textbf{Collection Style} & 0.00 & 0.00 & 0.00 & 0.00 & 0.00 & 1.59 & 0.00 \\
\textbf{Volume} & -1.59 & -4.76 & -2.38 & -0.79 & -4.76 & -1.59 & -3.17 \\
\textbf{Unit} & 0.00 & 0.79 & 0.79 & 0.00 & 0.00 & -0.79 & -0.79 \\
\textbf{Ethical Risks} & -0.79 & 0.79 & 0.00 & 0.00 & 0.00 & 0.00 & {1.59} \\
\textbf{Provider} & -1.59 & 0.00 & 0.00 & 0.00 & 0.00 & -0.79 & -0.79 \\
\textbf{Derived From} & -2.38 & -0.79 & 0.79 & -0.79 & 0.00 & -1.59 & -2.38 \\
\textbf{Tokenized} & {1.59} & 0.79 & 0.00 & {0.79} & 0.00 & 0.00 & -1.59 \\
\textbf{Host} & 0.00 & -0.79 & 0.00 & 0.00 & 0.00 & -1.59 & {1.59} \\
\textbf{Access} & -0.79 & -0.79 & -0.79 & -0.79 & 0.00 & -1.59 & -0.79 \\
\textbf{Cost} & -0.79 & -0.79 & 0.00 & 0.00 & 0.00 & 0.00 & 0.00 \\
\textbf{Test Split} & -0.79 & 0.79 & 0.00 & 0.00 & {0.79} & 0.79 & -0.79 \\
\textbf{Tasks} & 0.00 & {2.38} & 0.79 & 0.00 & 0.00 & {2.38} & {1.59} \\
\bottomrule
\end{tabular}
\end{table*}

\section{Full metrics}
As we explained in our data annotation procedure, we also add a binary value to indicate whether the attribute is actually extracted from the paper or elsewhere. To test the model's ability in only attributes that exist in papers, we show the results in Table \ref{tab:annotations}. As expected, considering only the attributes in the paper, we observe a significantly higher recall compared to precision.
\begin{table}[]
\centering
\small
\caption{Comparison between models in terms of precision, recall, and f1 scores. }
\label{tab:annotations}
\begin{tabular}{lccc}
\toprule
\textbf{Model} & \textbf{Precision} & \textbf{Recall} & \textbf{F1} \\
\midrule
\textbf{Random} & 29.46 & 30.27 & 29.74 \\
\textbf{Keyword} & 42.39 & 45.65 & 43.89 \\
\textbf{Gemma 3 27B} & 64.08 & 68.72 & 66.23 \\
\textbf{Qwen 2.5 72B} & 65.78 & 70.28 & 67.88 \\
\textbf{Llama 4 Maverick} & 67.04 & 71.90 & 69.30 \\
\textbf{DeepSeek V3} & 67.72 & 72.70 & 70.03 \\
\textbf{Claude 3.5 Sonnet} & \underline{69.08} & 73.94 & 71.34 \\
\textbf{GPT 4o} & 69.01 & \underline{74.09} & \underline{71.38} \\
\textbf{Gemini 2.5 Pro} & \textbf{72.67} & \textbf{78.17} & \textbf{75.23} \\
\bottomrule
\end{tabular}
\end{table}

\section{System Prompt}
In Figure \ref{fig:sys_prompt}, we show the system prompt to generate the metadata given the paper and the input schema. 
\begin{figure*}
    \centering
    \caption{System prompt for generating the metadata.}
    \label{fig:sys_prompt}
    \small
    \begin{tcolorbox}[colback=black!12, boxrule=0pt]
    You are a professional research paper reader. You will be provided 'Input schema' and 'Paper Text' and you must respond with an 'Output JSON'.
    The 'Output Schema' is a JSON with the following format key:answer where the answer represents an answer to the question. 
    The 'Input Schema' has the following main fields for each key:
    'question': A question that needs to be answered.
    'options' : If the 'question' has 'options' then the question can be answered by choosing one or more options depending on 'answer\_min' and 'answer\_max'
    'options\_description': A description of the 'options' that might be unclear. Use the descriptions to understand the options. 
    'answer\_type': the type of the answer to the 'question'. The answer must follow the type of the answer. 
    'answer\_min' : If the 'answer\_type' is List[str], then it defines the minimum number of list items in the answer. Otherwise it defines the minimum number of words in the answer.
    'answer\_max' : If the 'answer\_type' is List[str], then it defines the maximum number of list items in the answer. Otherwise it defines the maximum number of words in the answer.
    The answer must be the same type as 'answer\_type' and its length must be in the range ['answer\_min', 'answer\_max']. If answer\_min = answer\_max then the length of answer MUST be answer\_min. 
    The 'Output JSON' is a JSON that can be parsed using Python `json.load()`. USE double quotes "" not single quotes '' for the keys and values.
    The 'Output JSON' has ONLY the keys: '{columns}'. The value for each key is the answer to the 'question' that represents the same key in the 'Input Schema'.
    \end{tcolorbox}
\end{figure*}

\section{Individual Attribute Evaluation}
In Table \ref{tab:individual}, we highlight the results on individual attributes of metadata. We observe that most models struggle with Links and License in general. This could be attributed to missing such attributes in most papers. To observe the effect of browsing, we plot the difference between browsing and non-browsing models in Table \ref{tab:individual-diff}. Generally, we see the most gain in extracting the license attribute, which is mostly reported in repositories. However, we see a decrease in some other attributes. For example, the Volume is decreased because some papers report different numbers than what is actually released in the data. Additionally, the Link attribute is worse because it seems browsing forces the model to change the link based on what is in the repository. Interestingly, Gemma's results increase across multiple attributes without any decrease. 

\begin{table}[htp!]
\centering
\small
\begin{tabular}{l c}
\toprule
\textbf{Model} & \textbf{Average} \\
\midrule
\textbf{Llama 4 Maveric} & 54.17 \\
\textbf{DeepSeek V3} & 60.00 \\
\textbf{Qwen 2.5 72B} & 61.11 \\
\textbf{Claude 3.5 Sonnet} & 61.73 \\
\textbf{Gemma 3 27B} & 61.90 \\
\textbf{GPT 4o} & \underline{66.05} \\
\textbf{Gemini 2.5 Pro} & \textbf{73.33} \\
\bottomrule
\end{tabular}
\caption{Model performance by evaluating on papers published after the model release.}
\end{table}

\section{Contamination}
To consider contamination, we also evaluate the LLMs using papers after the model release. The results show no significant difference that might indicate the presence of some contamination.  

\section{Schema}
In Code \ref{lst:schema}, we show the full schema used for the Arabic datasets. Note in some experiments, we change the answer\_max to reflect to different constraints. 
\newpage
\onecolumn
\begin{lstlisting}[language=Python, caption=Example Schema for Arabic datasets' metadata, breaklines=true, numbers=none, label=lst:schema]
{
    "Name": {
        "question": "What is the name of the dataset?",
        "answer_type": "str",
        "answer_min": 1,
        "answer_max": 5
    },
    "Subsets": {
        "question": "What are the dialect subsets of this dataset? The keys are 'Name', 'Volume', 'Unit', 'Dialect'. 'Dialect' must be one of the country name from the dialect options",
        "answer_type": "List[Dict[Name, Volume, Unit, Dialect]]",
        "validation_group":"DIVERSITY",
        "answer_min": 0,
        "answer_max": 29
    },
    "Link": {
        "question": "What is the link to access the dataset? The link must contain the dataset. If the dataset is hosted on HuggingFace, use the HF Link.",
        "answer_type": "url",
        "validation_group":"ACCESSIBILITY",
        "answer_min": 1,
        "answer_max": 1
    },
    "HF Link": {
        "question": "What is the Huggingface link of the dataset?",
        "answer_type": "url",
        "validation_group":"ACCESSIBILITY",
        "answer_min": 0,
        "answer_max": 1
    },
    "License": {
        "question": "What is the license of the dataset?",
        "options": [
            "Apache-1.0",
            "Apache-2.0",
            "Non Commercial Use - ELRA END USER",
            "BSD",
            "CC BY 1.0",
            "CC BY 2.0",
            "CC BY 3.0",
            "CC BY 4.0",
            "CC BY-NC 1.0",
            "CC BY-NC 2.0",
            "CC BY-NC 3.0",
            "CC BY-NC 4.0",
            "CC BY-NC-ND 1.0",
            "CC BY-NC-ND 2.0",
            "CC BY-NC-ND 3.0",
            "CC BY-NC-ND 4.0",
            "CC BY-SA 1.0",
            "CC BY-SA 2.0",
            "CC BY-SA 3.0",
            "CC BY-SA 4.0",
            "CC BY-NC 1.0",
            "CC BY-NC 2.0",
            "CC BY-NC 3.0",
            "CC BY-NC 4.0",
            "CC BY-NC-SA 1.0",
            "CC BY-NC-SA 2.0",
            "CC BY-NC-SA 3.0",
            "CC BY-NC-SA 4.0",
            "CC0",
            "CDLA-Permissive-1.0",
            "CDLA-Permissive-2.0",
            "GPL-1.0",
            "GPL-2.0",
            "GPL-3.0",
            "LDC User Agreement",
            "LGPL-2.0",
            "LGPL-3.0",
            "MIT License",
            "ODbl-1.0",
            "MPL-1.0",
            "MPL-2.0",
            "ODC-By",
            "unknown",
            "custom"
        ],
        "answer_type": "str",
        "validation_group":"ACCESSIBILITY",
        "answer_min": 1,
        "answer_max": 1
    },
    "Year": {
        "question": "What year was the dataset published?",
        "answer_type": "date[year]",
        "answer_min": 1,
        "answer_max": 1
    },
    "Language": {
        "question": "What languages are in the dataset?",
        "options": ["ar", "multilingual"],
        "option_description": {
            "ar": "the dataset is purely in Arabic, there are no other languages involved",
            "multilingual": "the dataset contains samples in other languages"
        },
        "answer_type": "str",
        "validation_group":"DIVERSITY",
        "answer_min": 1,
        "answer_max": 1
    },
    "Dialect": {
        "question": "What is the dialect of the dataset?",
        "options": [
            "Classical Arabic",
            "Modern Standard Arabic",
            "United Arab Emirates",
            "Bahrain",
            "Djibouti",
            "Algeria",
            "Egypt",
            "Iraq",
            "Jordan",
            "Comoros",
            "Kuwait",
            "Lebanon",
            "Libya",
            "Morocco",
            "Mauritania",
            "Oman",
            "Palestine",
            "Qatar",
            "Saudi Arabia",
            "Sudan",
            "Somalia",
            "South Sudan",
            "Syria",
            "Tunisia",
            "Yemen",
            "Levant",
            "North Africa",
            "Gulf",
            "mixed"
        ],
        "option_description": {
            "mixed": "the dataset contains samples in multiple dialects i.e. social media. Assume Modern Standard Arabic if not specified."
        },
        "answer_type": "str",
        "validation_group":"DIVERSITY",
        "answer_min": 1,
        "answer_max": 1
    },
    "Domain": {
        "question": "What is the source of the dataset?",
        "options": [
            "social media",
            "news articles",
            "reviews",
            "commentary",
            "books",
            "wikipedia",
            "web pages",
            "public datasets",
            "TV Channels",
            "captions",
            "LLM",
            "other"
        ],
        "answer_type": "List[str]",
        "validation_group":"CONTENT",
        "answer_min": 1,
        "answer_max": 11
    },
    "Form": {
        "question":"What is the form of the data?", 
        "options": ["text", "spoken", "images"], 
        "answer_type": "str", 
        "validation_group":"CONTENT", 
        "answer_min": 1,
        "answer_max": 1
    },
    "Collection Style": {
        "question": "How was this dataset collected?",
        "options": [
            "crawling",
            "human annotation",
            "machine annotation",
            "manual curation",
            "LLM generated",
            "other"
        ],
        "option_description": {
            "crawling": "the dataset was collected by crawling the web",
            "human annotation": "the dataset was labelled by human annotators",
            "machine annotation": "the dataset was collected/labelled by machine programs",
            "manual curation": "the dataset was collected manually by human curators",
            "LLM generated": "the dataset was generated by an LLM",
            "other": "the dataset was collected in a different way"
        },
        "answer_type": "List[str]",
        "validation_group":"CONTENT",
        "answer_min": 1,
        "answer_max": 5
    },
    "Description": {
        "question": "Write a brief description about the dataset",
        "answer_type": "str",
        "answer_min": 0,
        "answer_max": 50
    },
    "Volume": {
        "question": "What is the size of the dataset?. If the dataset is multilingual only use the size of the Arabic dataset",
        "answer_type": "float",
        "validation_group":"CONTENT",
        "answer_min": 1
    },
    "Unit": {
        "question": "What kind of examples does the dataset include?",
        "options": ["tokens", "sentences", "documents", "hours", "images"],
        "option_description": {
            "tokens": "the dataset contains individual tokens/words",
            "sentences": "the samples are sentences or short paragraphs",
            "documents": "the samples are long documents i.e. web pages or books",
            "hours": "the samples are audio files",
            "images": "the samples are images"
        },
        "answer_type": "str",
        "validation_group":"CONTENT",
        "answer_min": 1,
        "answer_max": 1
    },
    "Ethical Risks": {
        "question": "What is the level of the ethical risks of the dataset?",
        "options": ["Low", "Medium", "High"],
        "option_description": {
            "Low": "most likely no ethical risks associated with this dataset",
            "Medium": "social media datasets",
            "High": "hate/offensive datasets from social media, or web pages"
        },
        "answer_type": "str",
        "validation_group":"CONTENT",
        "answer_min": 1,
        "answer_max": 1
    },
    "Provider": {
        "question": "What entity is the provider of the dataset? Don't use Team.",
        "answer_type": "List[str]",
        "validation_group":"ACCESSIBILITY",
        "answer_min": 0,
        "answer_max": 10
    },
    "Derived From": {
        "question": "What datasets were used to create the dataset?",
        "answer_type": "List[str]",
        "validation_group":"EVALUATION",
        "answer_min": 0
    },
    "Paper Title": {
        "question": "What is the title of the paper?",
        "answer_type": "str",
        "answer_min" : 3
    },
    "Paper Link": {
        "question": "What is the link to the paper?",
        "answer_type": "str",
        "answer_min": 1
    },
    "Script": {
        "question": "What is the script of this dataset?",
        "options": ["Arab", "Latin", "Arab-Latin"],
        "option_description": {
            "Arab": "The script used is only in Arabic",
            "Latin": "The script used is only in Latin i.e. it has samples written in Latin like Arabizi or transliteration",
            "Arab-Latin": "The script used is a mix of Arabic and Latin"
        },
        "answer_type": "str",
        "validation_group":"CONTENT",
        "answer_min": 1,
        "answer_max": 1
    },
    "Tokenized": {
        "question": "Is the dataset tokenized?",
        "options": [true, false],
        "option_description": {
            "true": "The dataset is tokenized. Tokenized means the words are split using a morphological analyzer",
            "false": "The dataset is not tokenized"
        },
        "answer_type": "bool",
        "validation_group":"CONTENT",
        "answer_min": 1,
        "answer_max": 1
    },
    "Host": {
        "question": "What is name of the repository that hosts the dataset?",
        "options": [
            "CAMeL Resources",
            "CodaLab",
            "data.world",
            "Dropbox",
            "Gdrive",
            "GitHub",
            "GitLab",
            "kaggle",
            "LDC",
            "MPDI",
            "Mendeley Data",
            "Mozilla",
            "OneDrive",
            "QCRI Resources",
            "ResearchGate",
            "sourceforge",
            "zenodo",
            "HuggingFace",
            "ELRA",
            "other"
        ],
        "answer_type": "str",
        "validation_group":"ACCESSIBILITY",
        "answer_min": 1,
        "answer_max": 1
    },
    "Access": {
        "question": "What is the accessibility of the dataset?",
        "options": ["Free", "Upon-Request", "With-Fee"],
        "option_description": {
            "Free": "the dataset is public and free to access",
            "Upon-Request": "the dataset is free to access but requires a submitting a request or filling out a form",
            "With-Fee": "the dataset is not free to access"
        },
        "answer_type": "str",
        "validation_group":"ACCESSIBILITY",
        "answer_min": 1,
        "answer_max": 1
    },
    "Cost": {
        "question": "If the dataset is not free, what is the cost?",
        "answer_type": "str",
        "validation_group":"ACCESSIBILITY",
        "answer_min": 0
    },
    "Test Split": {
        "question": "Does the dataset contain a train/valid and test split?",
        "options": [true, false],
        "option_description": {
            "true": "The dataset contains a train/valid and test split",
            "false": "The dataset does not contain a train/valid or test split"
        },
        "answer_type": "bool",
        "validation_group":"EVALUATION",
        "answer_min": 1,
        "answer_max": 1
    },
    "Tasks": {
        "question": "What NLP tasks is this dataset intended for?",
        "options": [
            "machine translation",
            "speech recognition",
            "sentiment analysis",
            "language modeling",
            "topic classification",
            "dialect identification",
            "text generation",
            "cross-lingual information retrieval",
            "named entity recognition",
            "question answering",
            "multiple choice question answering",
            "information retrieval",
            "part of speech tagging",
            "language identification",
            "summarization",
            "speaker identification",
            "transliteration",
            "morphological analysis",
            "offensive language detection",
            "review classification",
            "gender identification",
            "fake news detection",
            "dependency parsing",
            "irony detection",
            "meter classification",
            "natural language inference",
            "instruction tuning",
            "Linguistic acceptability",
            "other"
        ],
        "answer_type": "List[str]",
        "validation_group":"EVALUATION",
        "answer_min": 1,
        "answer_max": 5

    },
    "Venue Title": {
        "question": "What is the venue title of the published paper?",
        "answer_type": "str",
        "answer_min": 1
    },
    "Venue Type": {
        "question": "What is the venue type?",
        "options": ["conference", "workshop", "journal", "preprint"],
        "answer_type": "str",
        "answer_min": 1,
        "answer_max": 1
    },
    "Venue Name": {
        "question": "What is the full name of the venue that published the paper?",
        "answer_type": "str",
        "answer_min" : 0 
    },
    "Authors": {
        "question": "Who are the authors of the paper?",
        "answer_type": "List[str]",
        "answer_min": 1,
        "answer_max": 20
    },
    "Affiliations": {
        "question": "What are the affiliations of the authors?",
        "answer_type": "List[str]",
        "answer_min": 0,
        "answer_max": 20
    },
    "Abstract": {
        "question": "What is the abstract of the paper? replace any double quotes in the abstract by single quotes '",
        "answer_type": "str",
        "answer_min": 5
    }
}
\end{lstlisting}

\end{document}